# AD-BERT: Using Pre-trained contextualized embeddings to Predict the Progression from Mild Cognitive Impairment to Alzheimer's Disease

Chengsheng Mao[1], Jie Xu[2,3], Luke Rasmussen[1], Yikuan Li[1], Prakash Adekkanattu[3], Jennifer Pacheco[1], Borna Bonakdarpour[4], Robert Vassar[4], Guoqian Jiang[5], Fei Wang[3], Jyotishman Pathak[3], Yuan Luo[1]

[1] Department of Preventive Medicine, Feinberg School of Medicine, Northwestern University, Chicago, IL;

[2] Department of Health Outcomes and Biomedical Informatics, University of Florida, Gainesville, FL;

[3] Weill Cornell Medicine, New York, NY;

[4] Department of Neurology, Feinberg School of Medicine, Northwestern University, Chicago, IL;

[5] Mayo Clinic, Rochester, MN

## Abstract

**Objective:** We develop a deep learning framework based on the pre-trained Bidirectional Encoder Representations from Transformers (BERT) model using unstructured clinical notes from electronic health records (EHRs) to predict the risk of disease progression from Mild Cognitive Impairment (MCI) to Alzheimer's Disease (AD).

**Materials and Methods:** We identified 3657 patients diagnosed with MCI together with their progress notes from Northwestern Medicine Enterprise Data Warehouse (NMEDW) between 2000-2020. The progress notes no later than the first MCI diagnosis were used for the prediction. We first preprocessed the notes by deidentification, cleaning and splitting, and then pretrained a BERT model for AD (AD-BERT) based on the publicly available Bio+Clinical BERT on the preprocessed notes. The embeddings of all the sections of a patient's notes processed by AD-BERT were combined by MaxPooling to compute the probability of MCI-to-AD progression. For replication, we conducted a similar set of experiments on 2563 MCI patients identified at Weill Cornell Medicine (WCM) during the same timeframe.

**Results:** Compared with the 7 baseline models, the AD-BERT model achieved the best performance on both datasets, with Area Under receiver operating characteristic Curve (AUC) of 0.8170 and F1 score of 0.4178 on NMEDW dataset and AUC of 0.8830 and F1 score of 0.6836 on WCM dataset.

**Discussion:** Clinical notes related to AD could have different terminologies in linguistic characteristics from clinical text for other diseases or general text, motivating the need for disease domain-specific BERT models for AD. AD-BERT show its superiority in MCI-to-AD prediction to other machine learning models including BERT models for general clinical text. Also, compared to the traditional biomarkers related to AD, EHRs usually contain routinely collected clinical data that reflects real-world evidence, suggesting EHRs as a promising way for AD related research. This study validated the effectiveness of clinical notes for MCI-to-AD prediction.

**Conclusion:** We developed a deep learning framework using BERT models which provide an effective solution for prediction of MCI-to-AD progression using clinical note analysis.

## Introduction

Alzheimer's Disease (AD) is the most prevalent progressive neurological disorder, and is the sixth leading cause of death in the United States since 2021[1]. Approximately 5.8 million adults 65 years or older are living with AD currently, and by some estimates, that number will increase to 14 million adults by the year 2030[2]. However, unfortunately, nearly 75% of patients suffering from AD remain undiagnosed globally due to stigma and lack of awareness[3].

The latest diagnostics guidelines classify AD into three stages based on patients' clinical symptoms[4]: preclinical AD[5], mild cognitive impairment (MCI) due to AD[6], and AD dementia[7]. People with MCI, which is marked by symptoms of abnormal memory and/or other thinking problems, may or may not progress to AD dementia, which is severe enough to impair a person's ability to function independently. Predicting the risk of MCI-to-AD progression is critical for clinical prognostication, risk stratification and early intervention[8]. Various approaches have been attempted to estimate the likelihood of disease progression from MCI to AD dementia using heterogeneous data modalities, including cerebrospinal fluid (CSF) biomarkers[9], magnetic resonance imaging (MRI)[10], positron emission tomography (PET)[11], genetics data[12] and the combination of them[13]. In addition, clinical measures such as Mini-Mental State Examination (MMSE) and Alzheimer's Disease Assessment Scale-Cognition (ADAS-Cog) have also been used for the prediction of MCI-to-AD progression[14-16]. Compared to the previous approaches, routinely collected clinical data from EHRs reflect real-world evidence. Specifically, clinical notes in EHRs contain rich information, such as family history, laboratory measurements, treatments, and self-report scores (e.g., MMSE), which may be analyzed to model the risk of disease progression from MCI-to-AD. Recently, studies have explored information from Electronic Health Records (EHRs) for AD-related research[17-19]. In particular, Fouladvand et al.[19] predicted the progression from cognitively unimpaired (CU) to MCI using demographics, clinical notes, and self-reported information. Wang et al.[18] proposed a deep learning model to detect the evidence of cognitive decline from clinical notes. However, research is still limited in analyzing EHRs to study MCI-to-AD progression prediction.

Recently, the Bidirectional Encoder Representations from Transformers (BERT)[20] model that was pretrained with contextualized embeddings has shown promising results in many NLP tasks, and has been extended to BioBERT[21] and Bio+Clinical BERT (BC-BERT)[22] for biomedical text and clinical narratives, respectively. These pre-trained language models have been used for the prediction or identification of unplanned readmission[23], heart failure[24], acute kidney injury[25] and others from clinical notes. However, to the best of our knowledge, no prior study has applied a pre-trained model like BERT to the research of Alzheimer's disease. Clinical notes related to MCI or AD usually have different terminologies in linguistic characteristics from clinical text for other diseases or general text. For example, the phrase "memory loss" appears more frequently in MCI or AD related clinical notes than in general text. This motivating the need for specific BERT models for MCI-to-AD prediction.

Based on above motivations, i.e., research is still limited in analyzing EHRs to study MCI-to-AD progression prediction and specific BERT models are needed to analyze EHR for MCI-to-AD prediction, we developed a deep learning framework based on BERT for MCI-to-AD risk prediction using clinical notes from EHRs, and validated it on an independent dataset from a different institute. In addition, for the MCI-to-AD prediction, the case group that contains the MCI patients progressing to AD usually has much less samples than the control group that contains the MCI patients who did not progressing to AD, causing the class extremely imbalanced. In a batch-based deep learning framework, the extremely imbalanced classes may cause a batch containing only control samples and no case samples, thus, making the training process fail. To address the class imbalance issue, we designed a stratified batch sampler to ensure that all batches have an equal ratio between case and control samples.

**Related Work**

Natural language processing (NLP) methods are broadly applied for the development of predictive models using clinical notes extracted from EHRs[26], some used the words directly as features and conventional machine learning methods to predict the patient outcomes[27-30], some enhanced conventional machine learning methods with Unified Medical Language System (UMLS)[31-37], some leveraged deep learning methods (e.g., convolutional neural networks) for disease onset and outcome prediction[38-45], some aimed to improve interoperability of NLP systems by anchoring on common data models[46,47]. In the context of AD, Fouladvand et al.[19] predicted the progression from cognitively unimpaired (CU) to MCI using demographics, clinical notes, and self-reported information and the best performing model (in this case, Long Short-Term Memory (LSTM)) achieved an AUC of 0.75 and F1 Score of 0.46. Wang et al.[18] proposed a deep learning model to detect the evidence of cognitive decline from clinical notes 4 years preceding the patient's first diagnosis of MCI. The proposed recurrent neural network demonstrated optimal predictive power of both AUC and AUPRC over 0.9. However, as far as we know, no prior study has attempted to predict the progression from MCI to AD using clinical narratives.

In recent years, transformer-based models, e.g., BERT and Generative Pre-trained Transformer (GPT), have been pushing the state-of-the-art for heterogeneous NLP tasks including question answering, document classification, text generation and others. After observing the success of those models in the general domain, researchers released biomedical adapted variants, including BioBERT[21], ClinicalBERT[22], TCM-BERT[48] and BlueBERT[49], by pre-training the attention-based transformer blocks with large scale clinical and biomedical corpora. These transformer-based models also achieved breakthrough results in phenotyping and clinical predictive tasks using EHR data. For instance, Venkatakrishnan et al.[50] investigated the associations between patients' pre-existing conditions and short-/long- term COVID-19 complications using a transformer-based model. The curated pre-existing conditions can be further used to predict the complications of COVID-19. In another study, Mao et al.[25] achieved better results in the prediction of AKI by pre-training a language model from clinical notes initialized by BERT weights. In the diagnosis of AD, transformer-based models are frequently involved to distinguish between healthy and afflicted patients from their speech. Both audio and text components of speech can be represented by transformer-based models. However, to the best of our knowledge, no experiment has been done to examine the predictive power of transformer-based models when applied to AD-related clinical records.

**Materials and Methods**

**Study cohort**

We identified a cohort of patients with MCI from the Northwestern Medicine Enterprise Data Warehouse (NMEDW) and Weill Cornell Medicine (WCM) using ICD-9 (331.83) and ICD-10 (G31.84) codes. Those who progressed to AD (identified by ICD-9 [331.0] and ICD-10 [G30.*]) were considered as the case group, and the control group was defined as MCI patients who have *not* yet been diagnosed with AD. We identified 396 cases and 3261 controls from NMEDW and

548 cases and 2015 controls from WCM between 2000 and 2020. The exclusion criteria for NMEDW and WCM dataset are depicted in Figure 1. The patients having only 1 encounter may indicate that their medical records are not complete in our healthcare system. It would be safer to exclude them than to count them as negative/control samples. All progress notes before the first encounter when a patient was diagnosed with MCI were collected for the risk prediction.

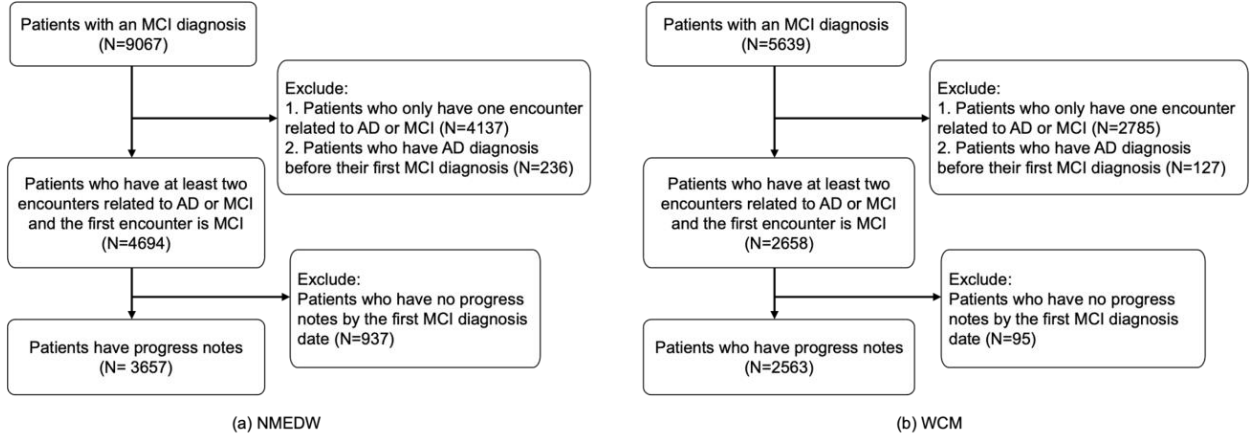

**Figure 1** Inclusion and exclusion criteria for the study cohorts for (a) NMEDW and (b) WCM.

**Preprocessing**

We preprocessed each progress note derived from the EHRs of all eligible patients as follows:
(1) Deidentification: Clinical notes contain legally Protected Health Information (PHI), such as patient names, addresses, and phone numbers, which should not be released to the public and should not be used for most research applications. We use the package Philter[51] to remove PHI for clinical notes.
(2) Cleaning: We removed non-ASCII characters from the notes and replaced multiple contiguous white spaces with one blank space.
(3) Splitting: Each note is split into sections by the newline character ('\n'). Each section is as an input to the model independently for section representation. And the embedding vector for a patient generated by the MaxPooling of all the section representations.

**Framework**

We first pretrained an AD-BERT model from BC-BERT[22] on the corpus of notes for all MCI patients from NMEDW dataset; the corpus contained about 37,000 clinical progress notes with an average length of 6250 characters. In the pretraining stage of AD-BERT, we followed the pretraining process of BERT with a loss function of the two unsupervised tasks, i.e., Masked Language Model (MLM) and Next Sentence Prediction (NSP), and pretrained AD-BERT from BC-BERT on the corpus of AD-related notes.

After AD-BERT was pretrained, we then fine-tuned AD-BERT for the MCI-to-AD prediction task using the training patients, and evaluated the model performance on an independent test set. The fine-tuning and testing were performed on both the NMEDW dataset and WCM dataset. Our framework is depicted in Figure 2, where all the encounter notes of a patient are split into sections that are then used as an input to the pretrained AD-BERT. Different patients can have different number of notes and sections; AD-BERT generates a section embedding for each section independently; all the section embeddings of a patient are then combined to generate a vector representation for that individual by MaxPooling. Then a fully-connected linear layer appended with a sigmoid activation is employed to predict the probability of MCI-to-AD progression.

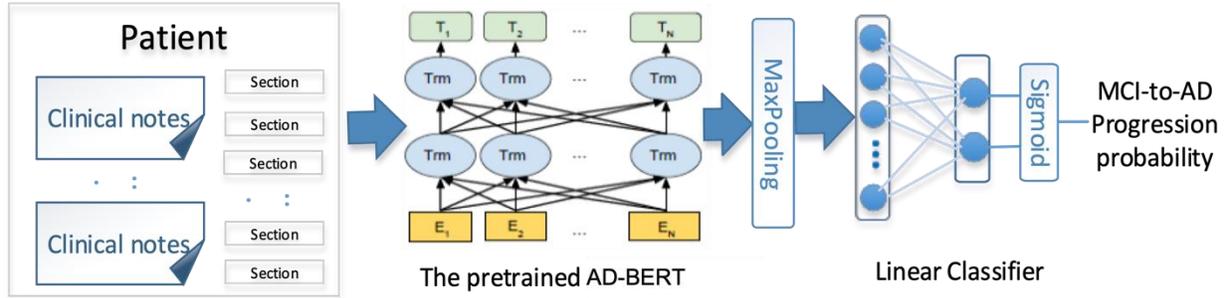

**Figure 2 The overview of our framework**

**Configuration**

We used the default configuration of BERT in *pytorch transformers*[52] for AD-BERT pretraining. For fine-tuning, each study cohort was randomly stratified into training and test set by 8:2, and the training set was further split 1/5 as the validation set that was used to select the best model in the training process.

Deep learning models are usually trained based on batches that only a batch of the training set is used to tune the parameters at a time by stochastic gradient decent[53]. Since the datasets were extreme imbalanced, i.e., we had much more control samples than case samples, model training based on random selected batches would be biased to the control even if we employed a class-weighted loss function, because batches generated by random selection were likely to contain no case samples. Using a large batch size might reduce the bias, but a large batch would cause the out-of-memory issue for model training, especially for long notes. To address this issue, we designed a stratified batch sampler to ensure the same case/control ratio in all batches, and used a class-weighted loss function based on the case/control ratio. In our study, the batch size was set to 4 and the max sequence length was set to 32. We used a weighted cross-entropy loss with the class weights inversely proportional to the corresponding number in a batch. WCM data was used to validate our framework and the pretrained AD-BERT on an independent data source.

**Results**

We identified a total of 3657 MCI patients from NMEDW and 2563 patients from WCM. The summary statistics of the patients for NMEDW and WCM are listed in Table 1 and Table 2, respectively. NMEDW dataset contained 396 MCI patients (233 [58.8%] females; age mean [SD] 76.8 [9.0] years) who had progressed to AD. WCM dataset contained 548 MCI patients (311 [56.8%] females; age mean [SD] 74.4 [6.7] years) who had progressed to AD. The average conversion time is 723 days and 742 days in NMEDW and WCM, respectively. The conversion time is defined by the time length from the first MCI diagnosis date to the first AD diagnosis date (based on ICD-9/10 codes: 331.0, G30.*) .

Besides the no-restrict prediction that is to predict whether an MCI patient will progress to AD without time restriction, we also considered the time windows of 6-month, 1-year, and 2-year to predict the MCI-to-AD progression, i.e., predicting whether an MCI patient would progress to AD in 6 months, 1 year, and 2 years, respectively. Samples in no-restrict prediction and *x*-month prediction are illustrated in Figure 3. For no-restrict prediction Figure 3(a), the case and control groups are distinct by the diagnosis condition, i.e., whether an AD diagnosis was found after MCI diagnosis. For a prediction in a certain time window of x months (x-month prediction) Figure 3(b), besides the diagnosis condition (i.e., whether an AD diagnosis was found in x months after MCI diagnosis), we also considered the time length condition for the control group, i.e., the last encounter should be found after x months to ensure the patients have a conversion time of at least x months. We excluded the patients whose last encounter is in the time window and no AD diagnosis is detected. The patient counts for each setting are shown in Table 3.

**Table 1 Summary statistics of patients in NMEDW dataset**

| Characteristic | Case (n=396) | Control (n=3261) |
|---|---|---|
| Age, Mean (SD), years | 76.8 (9.0) | 71.6 (15.8) |
| **Sex, No. (%)** | | |
| Female | 233 (58.8) | 1680 (51.5) |
| **Race, No. (%)** | | |
| White | 311 (78.5) | 2356 (72.2) |
| Black | 31 (7.8) | 401 (12.3) |
| Asian | 3 (0.8) | 79 (2.4) |
| Others & Unknown | 51 (12.9) | 425 (13.0) |
| Average conversion days | 723 (MCI to AD) | 521 (MCI lasts) |

**Table 2 Summary statistics of patients in WCM dataset**

| Characteristic | Case (n=548) | Control (n=2015) |
|---|---|---|
| Age, Mean (SD), years | 74.4 (6.7) | 69.3 (13.0) |
| **Sex, No. (%)** | | |
| Female | 311 (56.8) | 1089 (54.0) |
| **Race, No. (%)** | | |
| White | 325 (59.3) | 1038 (51.5) |
| Black | 28 (5.1) | 154 (7.7) |
| Asian | 9 (1.6) | 55 (2.7) |
| Others & Unknown | 186 (34.0) | 768 (38.1) |
| Average conversion days | 742 (MCI to AD) | 511 (MCI lasts) |

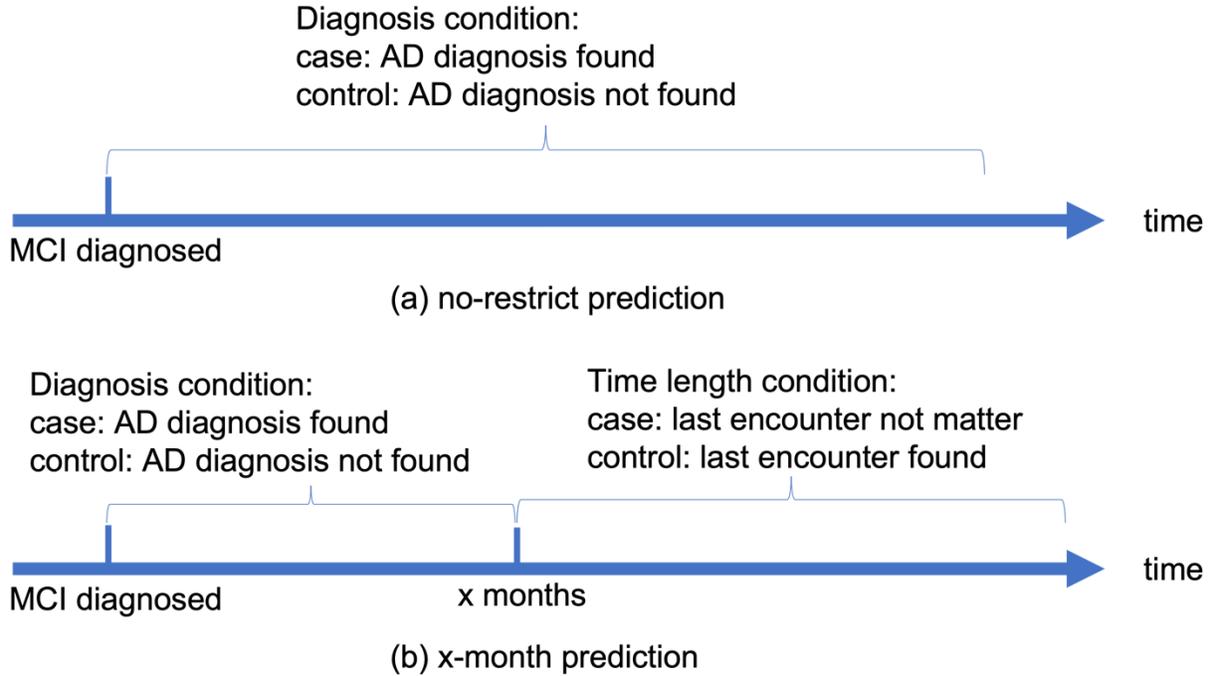

**Figure 3** The illustration of samples in (a) no-restrict prediction and (b) x-month prediction

Table 3 Patient counts in each setting in NMEDW and WCM

|  | NMEDW dataset | | | WCM dataset | | |
|---|---|---|---|---|---|---|
| Conversion time | case | control | total | case | control | total |
| 6 months | 156 | 1295 | 1451 | 141 | 1463 | 1064 |
| 1 year | 217 | 959 | 1176 | 221 | 1087 | 1308 |
| 2 years | 277 | 615 | 892 | 338 | 644 | 982 |
| no-restrict | 396 | 3261 | 3657 | 548 | 2015 | 2563 |

We compared our model with several popular models for text classification, including BERT-based models such as basic BERT[20], BC-BERT[22], and BioBERT[21], bidirectional LSTM[54,55] with or without attention, CNN[56] and BOW with logistic regression[28]. We evaluated the predictive ability of a predictive model based on two metrics, F1 score and AUC. The prediction results of our framework on the independent test set compared to the baseline models are shown in Table 4. On NMEDW dataset, among all the models, AD-BERT achieved the best performance for all the settings, including no-restrict, 6-month, 1-year and 2-year predictions; the AUC for no-restrict prediction is 0.8170 and the F1 score is 0.4178, representing an increase of 2% in F1 and 0.5% in AUC compared with the best baseline mode (BioBERT). The AD-BERT pretrained on NMEDW dataset also generalized well on the WCM dataset, achieving the best performance for all settings except 1-year prediction, with the no-restrict AUC 0.8830 and the F1 score 0.6836, representing an increase of 5.5% in F1 and 0.6% in AUC compared with the baseline model (BioBERT).

**Table 4 The performance of our model compared with other baselines for MCI to AD prediction. BiLSTM-att: bidirectional LSTM with attention. BOW+LR: bag-of-words and logistic regression.**

| Dataset | model | no-restrict | | 6-month | | 1-year | | 2-year | |
|---|---|---|---|---|---|---|---|---|---|
| | | F1 | AUC | F1 | AUC | F1 | AUC | F1 | AUC |
| NMEDW | AD-BERT (ours) | **0.4178** | **0.8170** | **0.3604** | **0.7715** | **0.4471** | **0.6967** | **0.5060** | **0.6817** |
| | BC-BERT | 0.3457 | 0.7826 | 0.2455 | 0.6330 | 0.4000 | 0.6467 | 0.5205 | 0.6785 |
| | BioBERT | 0.4096 | 0.8129 | 0.2621 | 0.6660 | 0.4314 | 0.6881 | 0.4078 | 0.5934 |
| | BERT | 0.3652 | 0.7825 | 0.2545 | 0.6330 | 0.3704 | 0.6170 | 0.4286 | 0.5446 |
| | BiLSTM | 0.3258 | 0.7212 | 0.2682 | 0.6612 | 0.3429 | 0.5986 | 0.4565 | 0.6407 |
| | BiLSTM-att | 0.3390 | 0.7491 | 0.2624 | 0.7151 | 0.3522 | 0.6092 | 0.4340 | 0.6152 |
| | CNN | 0.3333 | 0.7493 | 0.3130 | 0.6803 | 0.3905 | 0.6199 | 0.4421 | 0.6422 |
| | BOW+LR | 0.1997 | 0.4617 | 0.1538 | 0.4292 | 0.2198 | 0.4613 | 0.1818 | 0.4390 |
| WCM | AD-BERT (ours) | **0.6836** | **0.8830** | **0.3562** | **0.6716** | 0.2556 | 0.6733 | **0.5596** | **0.6404** |
| | BC-BERT | 0.4957 | 0.7439 | 0.2516 | 0.6659 | **0.3862** | **0.6950** | 0.3937 | 0.5435 |
| | BioBERT | 0.6481 | 0.8780 | 0.2000 | 0.5859 | 0.3741 | 0.6930 | 0.2273 | 0.5242 |
| | BERT | 0.6447 | 0.8778 | 0.2796 | 0.6713 | 0.3810 | 0.6710 | 0.3600 | 0.5772 |
| | BiLSTM | 0.4603 | 0.6816 | 0.1702 | 0.5075 | 0.3273 | 0.5700 | 0.5172 | 0.5506 |
| | BiLSTM-att | 0.4702 | 0.7466 | 0.2174 | 0.5707 | 0.3220 | 0.5906 | 0.4837 | 0.6001 |
| | CNN | 0.5017 | 0.7401 | 0.1905 | 0.5876 | 0.3382 | 0.6193 | 0.3846 | 0.5174 |
| | BOW+LR | 0.2500 | 0.5034 | 0.1596 | 0.4812 | 0.2169 | 0.4812 | 0.4878 | 0.5658 |

We also find that deep learning models (e.g., BERT-based models, LSTM and CNN) achieve higher performance than BOW, indicating that only word counts in the notes cannot provide any information for MCI-to-AD prediction. AD-BERT pretrained on NMEDW can also perform well on WCM dataset, validating the generalizability of our framework.

We also investigated the prediction process to see what words the model will take more attention to make the prediction. Figure 4 shows an example of attention visualization of AD-BERT. From Figure 4, AD-BERT pays more attention to the terms like "memory", "MCI" and "difficulty recalling dates" than others, which is reasonable as serious memory loss is a typical symptom of AD.

progress notes by <**phi**person**>, MD at <**phi**date**>. physician filed: <**phi**date**>. note time: <**phi**date**>. status: <**phi**person**>, MD (physician) neurobehavior and memory clinic consultation note <**phi**person**> is a 71 y/o former engineer with PMH remarkable for hypercholesterolemia here as a self referral for second opinion on memory loss he is accompanied by wife and daughter. he thinks he has more difficulty recalling dates. daughter states prior workup by neurologist has not been impressive and there is concern regarding vagueness of diagnosis (MCI vs dementia and subtype). memory difficulties began approximately 3 years ago and have increasingly become more notica ble to family members. they feel patient himself has good insight into limitations but attempts to cover them up. short term memory loss is a major issue. conversational repetitiveness has increased. he has been known to forgot events (both in entirety and in terms of details).

**Figure 4 Attention visualization of AD-BERT**

**Discussion**

Recently, data from EHRs is being increasingly analyzed for AD-related research[17-19], but research is limited in the application of NLP methods on clinical notes to study MCI-to-AD progression prediction.

Given that EHR is invasive and reflects real-world evidence compared to traditional biomarkers of AD, in this study, we developed a model to automatically predict the progression from MCI to AD using information and features derived from unstructured progress notes in EHR. Overall, our framework (AD-BERT) achieved the highest performance among commonly used NLP models. AD-BERT is a BERT model pretrained on AD-related clinical notes; it could effectively catch AD related information in the clinical notes; thus AD-BERT is more effective than the models without pretraining on AD-related notes for AD-related tasks such as MCI-to-AD prediction.

In prior work, Young et al.[57] and Davatzikos et al.[13] achieved AUCs of 0.795 and 0.734, respectively, in the prediction of MCI-to-AD conversion using multiple biomarkers including MRI, PET, CSF and APOE. Zhang et al.[10] achieved an AUC of 0.888 in distinguishing MCI converter and non-converter using sMRI and fMRI data. In comparison to these studies, we note that clinical notes are more routinely available for a larger sample of patients than biomarker data, and our performance also validates the effectiveness of using clinical notes for MCI-to-AD prediction.

Furthermore, in this study, besides the no-restrict prediction, we also performed the MCI-to-AD prediction in 6 months, 1 year and 2 years. From the results, in terms of F1 score, we found that the performance of the 2-year prediction model was the best, and whereas, that of the 6-month model was the worst. This was expected because the prediction of a longer time window is easier than that of a shorter time window.

While NLP and ML technologies are very promising, a major challenge in broader adoption of these technologies is their portability across multiple EHR systems - i.e. developing methods that yield consistent results when applying to multiple, diverse settings. Coding practices, clinical practice patterns and physician behavior vary between clinical settings and will be reflected in clinical notes created at these sites. While previous studies have demonstrated varying levels of success in portability of NLP/ML technologies across institutions, recent work by Carrell et al. argue that there remain significant challenges in adapting NLP systems across multiple sites, which include assembling clinical corpora, managing diverse document structures and handling idiosyncratic linguistic expressions[58]. The fact that the AD-BERT model pretrained on NMEDW also performed well on WCM dataset further supports the generalizability of the framework.

Another innovation of this paper is the use of a stratified batch sampler to address the problem of data imbalance. Since the datasets in our study are quite imbalanced, the random selected batches usually contained no case samples, making the model biased to the control class. In this paper, we developed a stratified batch sampler algorithm to ensure each batch contains an equal proportion of case samples. We also conducted the experiments using the random selected batches without stratified batch sampler, where all the BERT-based models output a 0 recall and NA precision for MCI-to-AD prediction, indicating that all the samples including case and control were predicted to the control group, making the model fail.

The study has several limitations. First, our model was pretrained on data from a single healthcare system (NMEDW) and validated using data from a single external dataset (WCM). Since clinical documentation and workflows vary significantly across healthcare systems, although our model pretrained on Northwestern data performed well on an external dataset (WCM), this does not guarantee the same level of performance on clinical notes from other healthcare systems. In our future work, additional validation studies will have to be conducted using clinical notes from multiple health care systems to pretrain a more generalized AD-BERT model.

Second, we only considered two stages (i.e., MCI and AD) in the AD progression process. AD progression is a complex long-term process, and our study to predict how likely an MCI patient will progress to AD represents a small fraction of AD progression. More effort will be made to find the risk factors that affect the AD progression in our future work, including normal to AD, normal to MCI, and progression between even more refined stages.

Third, the data usually has some records with missing values that may somewhat influence the truth. In our study, we excluded the patient encounters with missing ICD codes. This may exclude encounters with AD, resulting assigning a case sample to the control group. For the missing ICD codes, we can only improve the data quality in the data base to reduce the missing rate. We also excluded the patients who have no progress notes when or before they were diagnosed with MCI. This will reduce the overall sample size, not likely to affect the overall prediction results. Moreover, the patients with missing progress notes only accounted for a small fraction (~0.2 in NMEDW).

Finally, we only considered the clinical notes for MCI-to-AD prediction in this study. Given that the biomarker data was also effective for AD prediction tasks, in our future work, we will try to combine clinical notes and biomarker data and other structured data to design an even more powerful multimodal approach for AD prediction.

## Conclusion

Clinical notes contain rich information that may suggest disease progression. However, it is challenging to extract predictive information from unstructured notes. In this paper, we developed a deep learning framework based on BERT for MCI-to-AD risk prediction using clinical notes from EHRs, and validated it on an independent dataset from a different institute. In addition, we designed a stratified batch sampler to address the class imbalance issue between case and control. The deep learning framework using BERT models in this study may provide a solution for clinical note analysis for MCI-to-AD progression.

Acknowledgment. This study was funded in part by NIH grants R01GM105688.